\documentclass[11pt]{article}
\usepackage{isolatin1}
\usepackage{acl2005}
\usepackage{balance}
\usepackage{times}
\usepackage{epsfig}
\usepackage{latexsym}
\usepackage{balance}
\usepackage{eso-pic} 
\usepackage{graphicx} 
\AddToShipoutPicture{\put(0,750){\texttt{Published in the Proceedings of Cross-Language Knowledge Induction Workshop, 2005 Cluj-Napoca}}}
\title{WSD Using English-Spanish Aligned Phrases over Comparable Corpora}

\author{David Fernández-Amorós\\
  Departamento de Lenguajes y Sistemas Informáticos\\
  UNED, Madrid\\
  {\tt david@lsi.uned.es}}

\date{} 

\begin{document}
\maketitle
\begin{abstract}

  In  this  paper  we  describe a  WSD  experiment  based on bilingual
  English-Spanish comparable corpora  in which individual noun phrases
  have been identified  and aligned with their respective counterparts
  in the  other language. The  evaluation of  the experiment has  been
  carried out against SemCor.

 We  show  that, with   the  alignment  algorithm employed,  potential
 precision is high (74.3\%), however the coverage of the method is low
 (2.7\%), due to alignments being far  less frequent than we expected. 

 Contrary to our intuition,  precision does not rise consistently with
 the  number of  alignments.  The  coverage   is low  due to   several
 factors; there are important  domain   differences, and English   and
 Spanish  are too close  languages for  this  approach  to be able  to
 discriminate efficiently between  senses, rendering it unsuitable for
 WSD, although   the  method may   prove more  productive  in  machine
 translation.

\end{abstract}

\section{Introduction}

Word Sense Disambiguation (WSD)  could   be defined  as the task    of
assigning  the right  sense to   a word  in    context given  a  sense
inventory.   This is a  problem in artificial intelligence reported at
least  since the nineteen fifties. There  is general  consensus in that
although it is not a very interesting question in itself in many areas
(lexicography  being the  obvious exception)   deeper understanding of
lexical  ambiguity would greatly  help  to solve some  applications of
natural language processing    and   clarify new  ones  still   to  be
uncovered.

 Here  we present   a  WSD experiment  based  on bilingual
English-Spanish comparable  corpora of   news  collections  in   which
individual   noun phrases have been  identified  and  aligned to their
counterparts in the other corpus.  WordNet \cite{WordNet} is a lexical
database for English which includes a sense inventory among many other
things.   This sense inventory relies  in the \emph{synset} concept. A
synset is  a  synonym  set of  words  with a  particular meaning,  for
instance two synsets associated  with  different senses of  church are
\{church,   Christian  church,   Christianity\}   and   \{church,  church
  building\}.      An  extension    of       WordNet    is  EuroWordNet
\cite{EuroWordNet}.  EuroWordNet  has   a  very  similar structure  to
WordNet, but comprises several European  languages. In addition, there
are links between the concepts in  different languages. The evaluation
of the experiment has been   carried out against  SemCor \cite{SemCor}.
SemCor  is  a collection of  English  texts  which  has been  manually
annotated with WordNet senses and for this  reason has often been used
as a test collection for WSD algorithms.

In a  first  step, the noun phrases  obtained   from the  English news
articles corpus are searched for in SemCor. Next, we associate each of
theses phrases  with  the corresponding  aligned  phrases in  Spanish,
together with the observed alignment frequency in the news collections.
In this alignments,  there is usually a cognate  or at least, one word
which is a direct translation of the other,  but the rest of the words
in the phrase can give us a clue about the correct sense.

The most relevant factors to consider about this experiment with respect to previous research are the following:
\begin{itemize}

\item Parallel vs.  Comparable   corpora. Many WSD algorithms   use  a
  supervised approach that relies on manually tagged examples to learn
  a  classification algorithm. This    manual tagging is  very  costly,
  leading to what has been called the knowledge acquisition bottleneck.
  A  relatively popular  approach has  been  to  use   parallel texts to  extract
  knowledge  automatically. The problem with  parallel corpora is that
  it is   also very  scarce. Comparable corpora   offers  some of  the
  advantages  of parallel corpora with a  much higher availability but at
  the cost of obtaining inferior quality knowledge.

\item Phrase detection.   It  is not  straightforward  to detect  noun
  phrases in different languages. We don't  know how big the impact of
  errors in detection is for the accuracy of this  approach.

\item Phrase aligning. Again, the precision  of the alignments between
  phrases (about 73\%), might affect the performance of the system.

\item The domain problem. It  is well-known that extracting  knowledge
  in one domain and  trying to apply it in  another one is generally a
  bad idea.   Ideally one should use   the same domain  for both tasks,
  however, it  is unlikely for   large unrestricted domain comparable corpora  to be widely
  available in the near future.

\item It is generally accepted that  one important obstacle for WSD is
  that cross language linguistic  effort has traditionally  focused on
  bilingual dictionaries  and  the like, which work   at word level or
  higher and that a reliable cross language man-made tool at the sense
  level would  greatly  contribute to the   solution  of the  problem.
  Fortunately  such  resources   now  exist;  the  set  of
  interlingual indices in EuroWordNet is an example.

\end{itemize}

In the second  section we discuss previous work  in the field together
with  a motivating example.  In the  following section we describe the
experiment. In   the third  section   we  present the  evaluation  and
results, with several successful and  unsuccessful examples and in the
fourth section we draw our conclusions and suggest  future work.

\section{Previous work}

The  basic idea,  is similar to  the  approach in \cite{Gale-93} which
uses the   English-French parallel corpus  of  the  Canadian Hansards,
although the fundamental unit from which information is extracted
is not the word but  the noun phrase, much  less ambiguous in general.
It   allows   discarding the  senses     of individual  words  when
translating  with  a  bilingual dictionary.    Our  approach is  more
related to  the work in \cite{Dagan-91,Dagan-94}  which uses  pairs of
syntactically related words.

This idea that different  senses of the same  word often translate  to
different words in a second language was also an argument to suggest a
new method of evaluating   WSD systems in \cite{Resnik-99a}.  The paper
presents  a formula to calculate  the  relatedness of  two word senses
according to the translations to a second language. Another novelty is
the generalization of the method to several pairs of languages instead
of just one.

The noun   phrases in English and Spanish   have been taken  from work
described in    \cite{Penas-02}  and the  alignment   between  them is
explained  in   \cite{Ostenero-02a,Ostenero-02b}.     The    alignment
algorithm  used has the  advantage  that corpora  doesn't  have to  be
parallel, just comparable. The phrases were presented to Spanish human
evaluators  in the interactive track of  the  CLEF'02 competition. The
evaluators had to find documents in a  database relevant to a query in
English with the  aid of text fragments  in Spanish. Using the phrases
in Spanish aligned with  the phrases in the  English documents as  aid
fragments considerably outperformed  SYSTRAN automatic translations of
the documents. These good results motivated the crossover attempt to WSD.

For the sake of clarity,  we sketch the procedure followed
to create the dictionary of aligned phrases.

The CLEF collections used to extract the  phrases are a Spanish corpus
made of   1994 news  from Spanish news  agency EFE and an  English
corpus containing  articles published by Los Angeles Times also
in 1994.

The first step is to identify the noun phrases. For Spanish, words are
lemmatized and POS tagged. After that process, chunks of words fitting
the following pattern are automatically considered noun phrases.

(noun $\mid$ adjective) (noun $\mid$ adjective $\mid$ preposition $\mid$ determiner $\mid$ conjunction )* (noun $\mid$ adjective)

Only phrases with two or three open-class words are considered because
the amount of longer phrases that can be aligned rapidly decreases.
This process identified more than  twenty-seven million alleged noun phrases  in
the corpus.

As far as  English is concerned, each  word was assigned the prior most
likely POS tag. The  pattern for identifying patterns was  the same  as in
Spanish. More than nine million  noun phrases for English were identified
this way.

The  alignment has  been carried out   with a bilingual resource.  The
phrases with two open-class words are aligned  with two open-class word phrases in
the other language and  so on for  three open-class word phrases.  The other
constraint required  to align    is that each    open-class word  in  a  phrase
translates to an open-class word in  the candidate phrase in the other
language. The real  alignment algorithm is  somewhat more complicated
but the details can be found in the referenced articles.

It has  been   shown in these  articles   that  the precision  in recognizing  noun phrases is high.  The  precision of the alignments
has been  estimated  in excess  of  73\%,  and the  correction in  the
alignment correlates with the  absolute frequency of the phrases, that
is, an alignment between commoner phrases is more likely to be correct.

We illustrate  the idea   with the  following example\footnote{Adapted
  from \cite{Ostenero-02a}}:

We   want  to disambiguate \emph{issue},   which  can be translated in
Spanish  as:  \emph{asunto,   tema,   número,  emisión,  expedición,
  descendencia, publicar, emitir,  expedir, dar y promulgar}.  At this
point we detect that the context of the word indicates that it is part
of the phrase \emph{abortion issue}.  This phrase has been aligned  with  the
phrase in Spanish  \emph{tema del aborto}.  

If we were doing machine translation, we would  be satisfied with this
translation, however, in the framework of WSD we would like to discard
the senses   of   \emph{issue} not  corresponding  to  the \emph{tema}
translation.  Unfortunately,    WordNet structure  does   not permit
obtaining  that  information  easily. The key    then, is to  associate
individual  word senses to translations  in the  other language, going
one step further from the word to word translation.

\section{The experiment}

In this section we describe the experiment. We start with the resource
of  the bilingual phrases     and the interlingual  indices (ILI)   in
EuroWordNet and we perform WSD in the SemCor collection. 

The approach   we have taken   uses EuroWordNet  interlingual indices.
These indices map the  \emph{synsets} from one EuroWordNet language to
another so that, in the previous example, we could use them to look up
the synsets associated with \emph{issue} in Spanish and find out which
of them hold the word \emph{tema}.  One drawback with this approach is
that  EuroWordNet taxonomy is  linked in English  with the concepts in
WordNet-1.5, which is a little outdated.  We want  to apply our system
to the senses in WordNet-1.7. This will allow the system, to be tested
in the short term with the latest SENSEVAL \cite{Kilgarriff-98a} collections
and thus compared with state-of-the-art participating  systems. To
overcome this version conflict we use the mappings from  versions 1.5 to 1.6
and from 1.6 to 1.7 of WordNet developed in \cite{Daude-00,Daude-01}.

As we want to disambiguate a target word, we first look at the context
to determine if the word belongs to one of the phrases with alignments
in our knowledge base.  We construct a simple automaton to implement a
detection algorithm which  takes  into account inflectional  variants.
We create a  forest in which trees  have  words as  labels.  Each tree
contains all  the phrases beginning  with a  certain word.   For  each
possible  continuation of a phrase beginning  with that word we have a
child node with the corresponding label. Some nodes as also labeled as
acceptance nodes, marking  the end of  a legal phrase (although  there
might be longer phrases with the same  prefix).  To detect the phrases
in the text, a word is read and looked up in the heads of the trees to
check for a match (two  words match if  their lemmas are the same). If
there is a the next word is read on and  tried to match with the label
of one of the children  nodes, and so forth until  no more matches are
possible.  If an acceptance node has been traversed then a noun phrase
has been found and the unused portion of the input  is restored to the
input buffer in order for the search to  continue.  Longer phrases are
preferred over shorter ones in case several acceptance nodes have been
traversed.  This algorithm takes linear time to detect the phrases.

If the word  belongs to a noun   phrase, we traverse  the list  of the
phrases aligned  with it in Spanish.  For  every word in  each Spanish
phrase, we  look  for the  synsets associated  with their senses  and,
using the ILI,   their associated English synsets.   If  any  of these
synsets  contains the target word,   then  the corresponding sense  is
kept, otherwise it is discarded.

We have  carried out  this process for  all  alignments,  and  we have
possibly discarded some  senses of the  word.  This way,  we use these
comparable  corpora as  a resource  to create  a filter, since several
senses  may remain for every  word.  Coverage  is  expected to be low,
still, a high precision  at discarding wrong senses   would make it  a
worthwhile approach  since that would  encourage further research, possibly scaling up  to multiple
pairs of languages instead of one.

The  first step to     prepare    the experiment has  consisted     of
automatically re-annotating  SemCor.  SemCor has manual annotations in
SGML of lemma, part of speech, WordNet  sense and even compound words,
among others, but  of course the information  about our dictionary  of
phrases is not  included so we translated  it into XML form and added,
for  the  words    belonging  to  aligned    phrases,  one  attribute,
\emph{phrase}, which  indicates   the  detected  phrase, and  another,
\emph{alignments},  which   shows a  list    of admissible senses with
respect to the algorithm just described, along with the frequency with
which the corpora  allowed a particular alignment supporting  that sense.   This
information is important, because reliability of alignment is supposed
to directly depend on its frequency.

So, for instance, this SemCor fragment:

\footnotesize
\begin{verbatim}
<wf cmd="done" pos="NN" lemma="number" 
wnsn="2"  lexsn="1:23:00::">number</wf>

<wf cmd="ignore" pos="IN">of</wf>

<wf cmd="done" pos="NN" lemma="voter" 
wnsn="1"  lexsn="1:18:00::">voters</wf>
\end{verbatim}
\normalsize

Would now look like this:

\footnotesize
\begin{verbatim}
<wf alignments="number%1:07:00:: 51 
   number%1:10:00:: 51 number%1:10:01:: 51 
   number%1:10:02:: 51 number%1:10:03:: 51 
   number%1:10:04:: 51 number%1:10:05:: 51 
   number%1:23:00:: 51" cmd="done" 
   lemma="number" lexsn="1:23:00::" 
  phrase="number of voters" pos="NN" 
  wnsn="2">number</wf>

<wf cmd="ignore" pos="IN">of</wf>
<wf cmd="done" lemma="voter" 
lexsn="1:18:00::"    pos="NN" wnsn="1">
voters</wf>
\end{verbatim}
\normalsize

It  is interesting  to note  that \emph{number}  has eleven  senses in
WordNet-1.7, of   which now only  eight  are  equally  amenable to  be
chosen.

\section{Evaluation and results}

We  have  evaluated this  approach  against SemCor.  This  decision is
supported by the fact that it  is a test  collection whose size allows
drawing more representative conclusions than from other, smaller-sized
collections, such as those in  \textsc{SENSEVAL} \cite{Kilgarriff-98a}.

In the process  of re-tagging the collection,  out of the 192840 words
amenable  for disambiguation  in   brown-1  and brown-2  segments,  we
detected 10787 English phrases, which make  up for 5.6\% of the words.
This phrases have alignments in Spanish in  5290 cases, so we filtered
senses for this number of words, 2.74\% of the total.  Among them, the
right  sense has  remained unfiltered   in 3922 cases.   That is,  the
filtering process has a potential precision of 74.33\%.

One   example in which  the algorithm  doesn't  work as expected is in
disambiguating \emph{friend} in the phrase \emph{friend of mine}.  The
alignment in Spanish was \emph{conocido de  las minas} (which could be
translated  as \emph{acquaintance  of  the   mines}). There are    two
relevant  observations.   First of  all, the  Spanish  phrase probably
refers to a  well-known flamenco festival which is  a  proper noun and
should therefore    not  be aligned  with a    common one.  An  entity
recognition module, even one  as simple as considering initial capital
letters,  should  have ruled this alignment   out. Second, one  has to
wonder how high can the degree of overlapping between the news in both
collections be.

It is obvious that alignment techniques  need to be improved. However,
since  these two phrases only were  aligned once we  felt  the need to
test  the correlation  between frequency of   alignments and potential
precision susceptible of being achieved.

\begin{center}
\begin{figure}[htbp]
\input{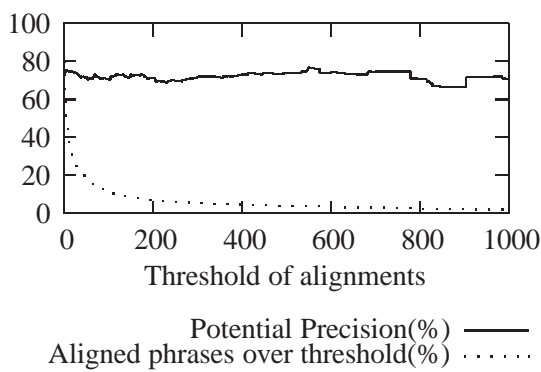}
    \caption{Relation between threshold, coverage and potential precision}
   \label{precumbral}
\end{figure}
\end{center}

In order to shed some light on the  subject we repeated the experiment
adding a threshold.  This time we  only disambiguate  words in phrases
having alignments with Spanish phrases when the alignment frequency is
over the threshold. Results can be seen in figure \ref{precumbral}.
The  results are  surprising:  Potential   precision does  not  really
increase    with  increasing      threshold  values.   Up    to   3000
occurrences\footnote{Not  shown in   the  graphic due  to the   almost
  negligible percentage   of phrases aligned},    there is hardly  any
difference in  potential precision. From there,  the number of phrases
with alignments is  so low (with  a threshold of 3000 occurrences  the
number of  phrases  with alignments  in SemCor  is just   38) that the
information is useless. Potential  precision of 100\% from a threshold
value of 8713 until the end (8836) corresponds solely to alignments of
the phrase \emph{year old}, which  equally support all four senses of
\emph{year}, so, at that point the information coming from the aligned
phrases is totally irrelevant.

As we can see,  the coverage of  the  approach is rather low,  but the
method really works, even when alignments  are only of modest quality,
to say the  least.  As a remarkable example,  the noun phrase
\emph{head of  the  family} was aligned  eight times   with the phrase
\emph{responsable de la cámara}. \emph{Head}  has 132 different senses
as  a   noun  in   WordNet-1.7.     \emph{Cámara} doesn't  help    but
\emph{responsable}  has two senses. One of   them is \{autor, culpable,
  perpetrador, responsable\} which the  ILI links to the English synset
\{culprit, perpetrator\}, which doesn't support any of the many senses of
\emph{head}.    However,    the  other    sense  of \emph{responsable}
corresponds to the synset \{responsable\}, linked  by ILI to the English
synset \{head, chief, top dog\} supporting the correct  sense of head in
the original phrase.

The mappings between WordNet versions and the Interlingual indices get
sometimes in the way  of the success. For the  sake of clarity we will
use the sense notation in  WordNet instead of  the associated synsets. The
sense notation refers to a set of lexicographer \emph{tematic} files. For
an illustration of  both problems in one  example, consider the phrase
\emph{art studies}. It is aligned with \emph{estudios de arte} (again,
a reasonably related phrase, but  highly unsuitable as a translation).
\emph{Arte} has four  senses: arte\%1:04:00::  which goes through  ILI plus 
the  mappings to  art\%1:04:00::, a  second one arte\%1:06:00::  which
points to   art\%1:06:00::,  a third  one,   arte\%1:09:00:: which ILI
points to  art\%1:09:00:: but the mappings just  don't map to anything
at all,  and the last sense  which  starts out as  arte\%1:10:00:: and
ends up as art\%1:10:00:: as expected.

The   mappings are not   complete  and therefore  occasionally fail to
upgrade a sense  to the  newer version,  but it is  more disturbing to
verify  that  \emph{art} and   \emph{arte} are  given  the exact  same
semantic   structure  in  EuroWordNet   as far    as the  algorithm is
concerned.

\section{Conclusions and future work}

We had hoped that use of  comparable corpora would help alleviate
the knowledge acquisition   bottleneck.  The corpora,  according to the
millions   of noun phrases detected,    seemed indeed bigger than many
parallel corpora available, and the scaling possibilities are obvious,
just  add    more years to  the    news collections.  Nevertheless, the
scarceness  of alignments has produced   an extremely low coverage for
the WSD  algorithm.  It is thus, very   unclear that bigger comparable
corpora would help WSD in this specific approach.

The domain problem has undeniably and heavily affected the experiment.
The Brown Corpus of which SemCor is a portion, was compiled from texts
printed   in 1961.  The  news  thirty-four  years  later surely  cover
different topics, many of which didn't even exist back in 1961. One of
the alignment   examples cited in  \cite{Ostenero-02b}  is  \emph{free
  trade agreement} which aligns with \emph{tratado de libre comercio}.
Even worse is that the noun phrases present in  SemCor hardly occur in
the  LAT '94  collection.  This  is  not a problem  of the noun-phrase
approach, but a  serious domain problem.   The question is that SemCor
is big  enough to allow interesting conclusions  to be  extracted from
experiments as  far  as statistics  are  concerned, but  very old with
respect   to  modern  texts.  There are    more  recent hand annotated
collections, however they are   much smaller-sized and thus  unfit for
statistically relevant purposes.

Also, regarding domain, it is  reasonable to suspect that the
differences between the news domain used to gather the phrases and the
SemCor collection, a part of the Brown Corpus, which was collected  with
the aim of being domain-free, might have influenced the results.

Another interesting  question regarding  coverage  is how   comparable
comparable corpora  really are. Ostenero  reports  38\% of the English
two-open-class-word noun-phrases to  have been aligned to Spanish ones
so the corpora seem moderately comparable.

The mappings used  to convert WordNet-1.5  synsets to version 1.7 have
been assessed by their authors to have  a precision around 90\%. Since
we   have   applied two  mappings   (1.5   $\rightarrow$  1.6 and  1.6
$\rightarrow$ 1.7)  we   can  estimate the probability   of  correctly
mapping a sense as being .9*.9=.81. That would account for 19\% of the
senses \emph{disappearing}  in the process, so  the real  reduction of
ambiguity  due to the  bilingual noun-phrase approach  has to be lower
than the overall figures apparently indicate.

The ILI has   proven to be somewhat   disappointing. In spite of  being
heavily advertised as one the of the  most outstanding achievements in
EuroWordNet,   it  turns  out    that  the   \emph{language   neutral}
representation  of nominal   entries to  which  the  ILI point  is
precisely the  nominal structure of  the original English WordNet-1.5.
Moreover,  in  the Spanish nominal   structure  the  contents of   the
synonyms sets  differ from the English ones  but the network structure
of hypernyms and other relations is the same except  when there is no
equivalent concept in Spanish or  the obvious linking is inapplicable.
That, plus the fact the  the ILI is semi-automatically constructed and
only   manually revised, amounts to    the English and Spanish nominal
structures being  so close that the ILI  coincides more often than desirable
with the identity function. This fact  is quite clear in the sense file
notation,  although the synset-offset number notation provides a rather
awkward encoding for this approximation of the function f(x)=x. This may
not be an issue with pairs of languages other than English but in this
case is a factor that requires further research.

The phrase detection  algorithm is in  its first stages of development
and there is much room for improvement, although it  is unknown if such
improvements will effectively help WSD.

The alignment technique employed  is also not  exempt of problems. The
algorithm seems very sound with respect  to finding correct alignments,
although we  suspect  that there is   a considerable  amount of  false
positives. If this problem was solved, potential precision could raise
a bit,  however it would definitely  lower a coverage that  is already
rather  tiny.   On  the  other   hand   there   are  cases   in  which
less-than-spectacular quality  alignments  have proven useful  for the
task.

This series of facts lead us  to conclude, in the  first place, that
although this method constitutes an \emph{a priori} interesting filter
in terms of precision, the rather low coverage  of the method produces
nearly negligible results for WSD.

Apart from   that  conclusion, the  most  interesting  result is that,
contrary  to our  intuition,    potential  precision does  not    rise
consistently with the number of alignments. Since the precision of the
alignments has  been  shown to correlate  with  the frequency  of  such
alignments,     the only explanation    is  that  these high-frequency
alignments are  not  productive in terms  of  filtering senses due  to
exactly equal mapping of  senses to words  in the  two languages.   We
observed this  behaviour in   the case  of   the most  aligned  phrase,
\emph{year old}.

So, in the  case of English and Spanish,  it was easy  to predict that
there  would be many   pathological cases.  For instance, the  phrases
containing   the word   \emph{art}  in   English  align  with  phrases
containing \emph{arte} in Spanish, something  which is not  productive
at all, since  all the senses of art  can be translated as \emph{arte}
and so  the  method does  not discard  any senses.  Alignments between
more heterogeneous pairs of  languages may improve performance as well
as adding  together  the results for  comparable  corpora  in multiple
pairs  of languages.  That  would  take  advantage of the   reasonable
hypothesis that ambiguities  will be different across different  pairs
of    languages.  The method may   prove   more productive in  machine
translation,   where many different  word  senses may translate to the
same  word in  the target language.  Of course,  the alignments are not
directly acceptable translations.

The  potential precision concept used for  the evaluation is certainly
somewhat fuzzy,  in that reduction    of ambiguity is  not  specified.
Potential  precision  is  not  to   be  confronted with   \emph{actual
  precision}, since this  approach only aspires to efficiently discard
some  senses of  the words, not  to  perform full disambiguation.
Anyway, the low  coverage of the method  allows to discard it  for WSD
purposes whatever the actual ambiguity reduction obtained.

These aligning   techniques were  successfully applied for    the CLEF
competition   \cite{Ostenero-02a},      in  human-computer interaction
scenario, however this success  does not carry  over to the  automated
WSD problem.  It is interesting to  note, however,  that since in that
work  the  phrases were   detected, aligned  and    used on  the  same
collections, there  were no domain  problems, thus obtaining much higher
coverage.

Summing up, the logical future work, in order for this approach to WSD
to  reach   viable status, would ideally comprise, among other
things,  finding  large quantities of   moderately parallel corpora in
different pairs of  languages (with  one of  them  fixed as target
language), of genre and age as close to those of the test collection as
possible, preferably  without    any  intervening    mappings.     The
predictable  much higher coverage of  the method would then foster the
need to measure  the  actual  degree of reduction   in  ambiguity.  The
question on the usefulness of the ILI remains open.

\section*{Acknowledgements}
We are indebted  to  Julio Gonzalo   for coming  up with  the  idea of
applying the noun phrases plus the  ILI to WSD and  for his advice, and
to  Fernando Lopez  Ostenero for his willing assistance  with  the
bilingual noun phrase resource.

\bibliographystyle{acl}
\bibliography{bibliografia}
\balance
\end{document}